\documentclass[conference,12pt]{article}

\usepackage[english]{babel}

\usepackage[a4paper, margin=1in]{geometry}

\usepackage{amsmath}
\usepackage[utf8]{inputenc}
\usepackage{graphicx}
\usepackage{subcaption}
\usepackage{amssymb}
\usepackage{float}
\usepackage{cite}
\usepackage[parfill]{parskip}

\usepackage{hyperref}
\hypersetup{colorlinks,allcolors=blue}

\begin{document}

\begin{center}
    {\Large\bfseries Deep Learning for Leopard Individual Identification:\\ An Adaptive Angular Margin Approach\par}
    \vspace{1em}
    David Colomer Matachana\\
    Imperial College London, Department of Earth Science \& Engineering\\
    \texttt{dc1823@ic.ac.uk}\\
    \vspace{2em}
\end{center}

\date{}

\begin{abstract}
Accurate identification of individual leopards across camera trap images is critical for population monitoring and ecological studies. This paper introduces a deep learning framework to distinguish between individual leopards based on their unique spot patterns. This approach employs a novel adaptive angular margin method in the form of a modified CosFace architecture. In addition, I propose a preprocessing pipeline that combines RGB channels with an edge detection channel to underscore the critical features learned by the model.

This approach significantly outperforms the Triplet Network baseline, achieving a Dynamic Top-5 Average Precision of 0.8814 and a Top-5 Rank Match Detection of 0.9533, demonstrating its potential for open-set learning in wildlife identification. While not surpassing the performance of the SIFT-based Hotspotter algorithm, this method represents a substantial advancement in applying deep learning to patterned wildlife identification.

This research contributes to the field of computer vision and provides a valuable tool for biologists aiming to study and protect leopard populations. It also serves as a stepping stone for applying the power of deep learning in Capture-Recapture studies for other patterned species.
\end{abstract}

\section{Introduction}

Possibly the most critical task in conservation consists of being able to identify individual animals over time. Through Capture-Recapture techniques, researchers can evaluate their fitness and track their demographics \cite{capture-recapture}. For example, in the case of leopards, monitoring their populations in an ever-growing urban landscape in India is the focus of intensive research \cite{Gubbi2020}, \cite{Gubbi}. 

Traditionally, this process has been carried out using invasive, telemetry-based methods; however, these techniques raise survival and stress concerns \cite{danger-invasive}, which pose significant issues when studying endangered species. With the advent of automatic camera traps, Photographic-Capture-Recapture (PCR) has gained popularity among researchers. This method relies on clear, distinct markings on the animals to facilitate individual identification. 

In earlier applications, the pattern matching was done visually by researchers \cite{manual-cap-recap}. This approach becomes extremely cumbersome for large datasets with thousands of images due to the need to attempt to match every pattern with every identified individual in the dataset. Moreover, it can lead to identification errors, notably the duplication of individuals, as it was seen in \cite{reduction-identification} that humans identified up to 22\% more cheetah individuals than the actual population. 

Given these limitations, it immediately became obvious that there was a need for semi-automatic pattern matching through Computer Vision (CV) algorithms. Today, there are numerous such programs. The state-of-the-art programs include Hotspotter \cite{Hotspotter} and Wild-ID \cite{WildID}. 

These algorithms are based on Scale Invariant Feature Transform (SIFT) \cite{Lowe2004} for explicit pattern extraction, and then use various distance-based or Machine Learning (ML) pattern matching algorithms. They offer varying results according to the species. I have not found studies on my target species, although, in a species with similar rosettes, the Jaguar, the accuracy reported is over 70\% \cite{sift2}. In other patterned species, accuracy reported ranges from 36\% \cite{Morrison2016} to close to 100\% \cite{Toad-Hotspotter}, with Hotspotter being the best performing software in general. Additionally, these systems require manual intervention to extract a Region of Interest (ROI) where the animal is located. This manual ROI extraction is not only time-consuming for large datasets but also introduces a layer of subjectivity and potential inconsistency.

Despite the efficiency of these algorithms, they were developed before recent advances in Deep Learning (DL) and may not harness the full potential of current technological capabilities. These methods can potentially achieve higher accuracy in individual animal identification due to their ability to learn complex patterns and features from large datasets implicitly. 

There have been some recent attempts to apply DL to this problem, notably to elephants \cite{deep8} and Giant Pandas \cite{deep5}. While they report high accuracies in re-identification, they have all built a traditional closed-set classifier, which means that their applicability outside of the training population is very limited. For example, \cite{deep5} report that their accuracy on a different population from the training descends from 95\% to 21\%. 

Methods suitable for open-set verification have been tested on Tigers \cite{TripletTiger}, but their applicability to diverse camera-trap images comes into question. This is due to the homogeneity of the training and test data for individuals, which consists of screenshots taken seconds apart that lack diversity in lighting, background, and pose for each individual. Moreover, the effectiveness of newer angular margin methods, such as CosFace \cite{Cosface}, has not been explored for patterned species individual identification.  

This paper attempts to address these gaps by building an end-to-end deep learning solution to the PCR problem on an open set. I propose a novel preprocessing pipeline that combines RGB channels with an edge detection channel to enhance the discriminative features learned by DL models. I develop both a Triplet Network with semi-hard negative mining and a novel adaptive angular margin CosFace architecture for leopard individual identification.

The objectives are to: 

\begin{enumerate}
    \item Evaluate whether recent advances in deep learning, specifically the modified CosFace architecture, can improve accuracy compared to the Triplet Network baseline for leopard identification.
    \item Assess if the proposed preprocessing technique can enhance the efficiency and accuracy of pattern extraction for individual leopard identification.
    \item Determine if the developed approach can achieve comparable performance to the SIFT-based Hotspotter system, potentially establishing it as a valid technique for PCR problems.
\end{enumerate}

\section{Methodology}

\subsection{Data Collection}

Diverse data is paramount to building an effective model. In this case, this becomes even more crucial, as fixed camera trap images of leopards are extremely variable in pose, lighting, etc.

To train the model, the \href{https://www.ncf-india.org/}{Nature Conservation Foundation} (NCF) has provided us with 8900 tagged images of 600+ individual leopards. Although the majority of images are of high quality, all images were inspected manually, of which 623 were removed given the impossibility of identifying an individual.

The remaining images have a skewed distribution in the number of images per leopard flank, with a mean of just 6.4 images per flank. While this dataset is quite extensive given the complexity of photographing and identifying such a low-density large carnivore, this is still orders of magnitude smaller than datasets used in facial recognition models, and thus poses challenges to reach comparable accuracies.

\begin{figure}[H]
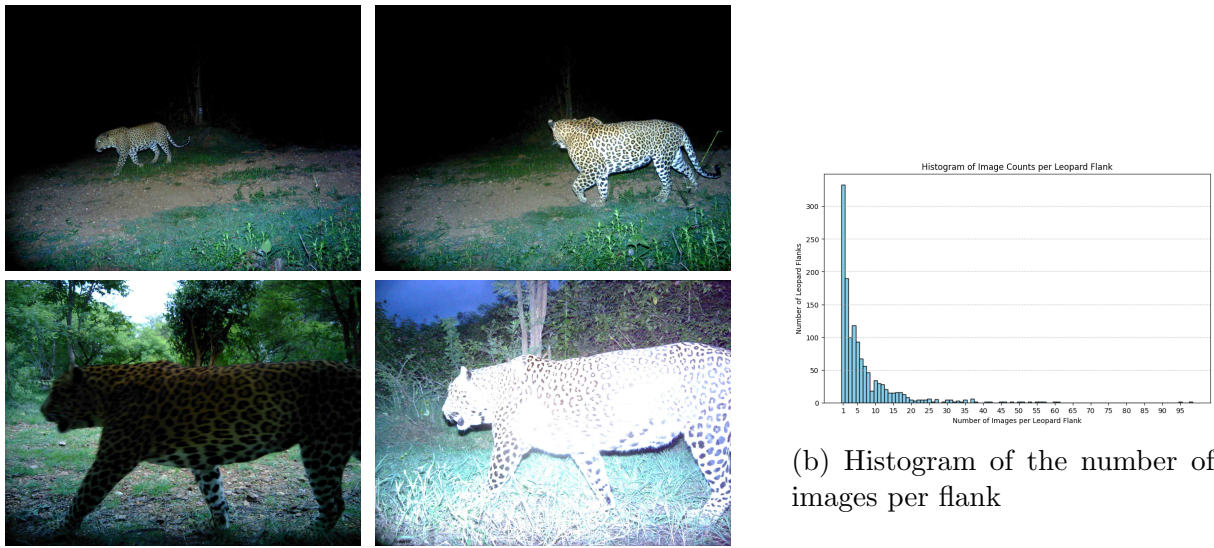

    \centering
    \begin{minipage}[c]{0.6\textwidth}
        \begin{subfigure}[c]{\textwidth}
            \centering
            \begin{minipage}[b]{0.49\linewidth}
                \centering
                \includegraphics[width=\textwidth]{BG-28B-2020-08-06_04-21-54.pdf}
            \end{minipage}
            \hfill
            \begin{minipage}[b]{0.49\linewidth}
                \centering
                \includegraphics[width=\textwidth]{BG-28B-2020-08-15_05-01-07.pdf}
            \end{minipage}
            
            \vspace{0.1cm}
            
            \begin{minipage}[b]{0.49\linewidth}
                \centering
                \includegraphics[width=\textwidth]{BG-36B-2020-08-02_15-15-08.pdf}
            \end{minipage}
            \hfill
            \begin{minipage}[b]{0.49\linewidth}
                \centering
                \includegraphics[width=\textwidth]{BG-42B-2020-08-08_05-57-41.pdf}
            \end{minipage}
            \caption{Comparison of 4 images of the same leopard among widely different poses, distances from the camera, and lighting conditions}
            \label{fig:leopard_grid}
        \end{subfigure}
    \end{minipage}%
    \hfill
    \begin{minipage}[c]{0.35\textwidth}
        \begin{subfigure}[c]{\textwidth}
            \centering
            \includegraphics[width=\textwidth]{histogram-images-per-leopard.pdf}
            \caption{Histogram of the number of images per flank}
            \label{fig:histogram}
        \end{subfigure}
    \end{minipage}
    \caption{(a) Four images of a leopard in different conditions. (b) Distribution of images per leopard flank.}
    \label{fig:combined}
\end{figure}

\subsection{Preprocessing}
\label{subsec: Preprocessing}
Given the large diversity in background, lighting, and most importantly distance of the subject from the camera, preprocessing is essential. The following preprocessing steps are built and then tested for effectiveness in enhancing the performance of the model.

\subsubsection{Bounding-box extraction}
Differing from Hotspotter and Wild-ID, in which the user has to manually identify the subject, I attempt to use current advances in CV for automatic detection extraction of bounding boxes, using a pre-trained YOLO network \cite{YOLO} for this, fine-tuned on camera trap images of animals \cite{PytorchWildlife}. The Network exhibits perfect accuracy in identifying all the usable leopard flanks. The only practical limitation observed was the tendency to occasionally detect other animals present in the images that were not leopards. These were manually removed.

\subsubsection{Background removal}
Furthermore, as seen in \cite{sift1}, filtering out the background results in a performance increase in previous re-identification algorithms. I thus use a background removal algorithm, "rembg" \cite{gatis2023rembg}. Unfortunately, even though the accuracy is still remarkable, it became clear that some usable leopard flanks are not accurately identified (3.2\% of the cases, most of them of poor quality), and some other flanks are just partially extracted (5.2\% of the cases). See figure \ref{fig:bad background} for examples.

\subsubsection{Edge Detection}
Edge-detection techniques as a preprocessing step have been associated with a performance increase for other classifiers in ML \cite{EdgeDetection}. However, to my knowledge, they still haven't been used in DL techniques for facial recognition. By isolating the leopard's rosette patterns and being invariant to lighting conditions, this approach aims to facilitate feature learning by the network. Histogram equalisation is first applied to the images before extracting the edges to ensure that all rosettes are considered as equal as possible. Light Gaussian Blurring is then used to only extract real edges, followed by Canny-Edge-Detection to isolate the patterns.

Although the vast majority of coats are fully extracted, 7.7\% of the images have partially extracted rosette patterns (see figure \ref{fig:edge detection}), primarily when a leopard is unevenly illuminated. These partial extractions may impact the model's performance. To mitigate this, the edge detected channel is concatenated with the other RGB channels.

\begin{figure}[ht]
    \centering
    \includegraphics[width=0.8\textwidth]{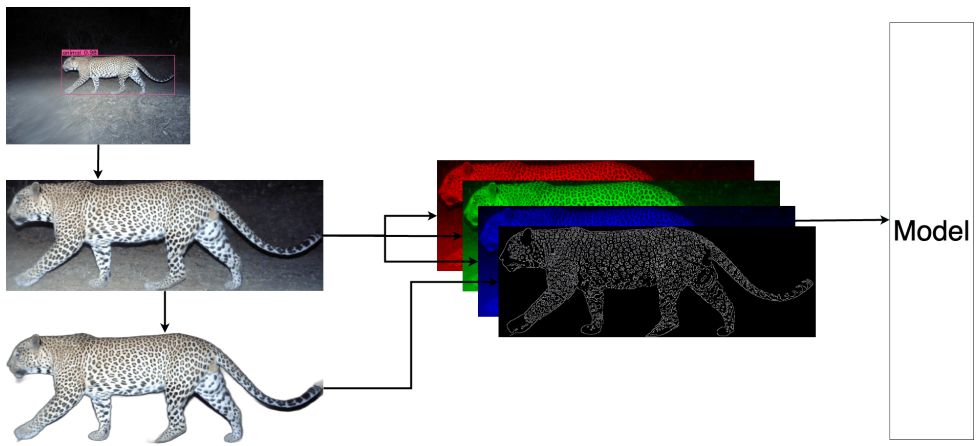}
    \caption{Full Preprocessing pipeline. Extract bounding box, remove background, edge detection. Input cropped RGB and edge detection to the model.}
    \label{fig:singleimage}
\end{figure}

\subsection{Architecture}
Open-set identification tasks pose significant issues to traditional classifiers. In these scenarios, the individuals seen during training differ from those in the application phase. Training traditional classifiers for this task is unsuitable, as the model learns to discriminate only between classes seen in training and cannot extrapolate to new individuals \cite{deep5}. 

Given this, inspiration is drawn from the parallels of this task with facial recognition models to design three architectures:

\newpage

\subsubsection{Triplet Convolutional Neural Network (Triplet CNN)}
\begin{figure}[ht]
    \centering
    \includegraphics[width=0.9\textwidth]{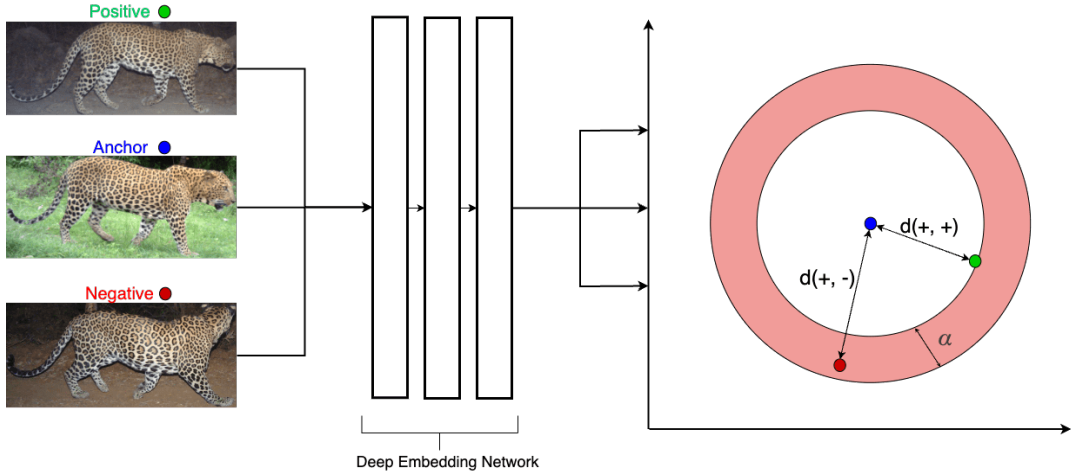}
    \caption{Triplet Network in training phase with Semi-hard negative mining: Negative embedding selected must be in the red area to satisfy condition. In the testing/inference phase, embeddings produced by the network are directly compared.}
    \label{fig:Triplet-architecture}
\end{figure}

In such an architecture, instead of outputting a class label, the network learns a feature space where embeddings of images are directly compared. The network is designed to process three images at a time: an anchor image, a positive image (i.e., another image of the same leopard), and a negative image (i.e., an image of a different leopard).

The Triplet CNN \cite{Triplet-Network}, \cite{architecture4} consists of three identical sub-networks sharing the same weights. Each sub-network outputs an embedding vector for its input image, and the similarity between these vectors is measured using the euclidean distance function.

The training objective is to minimise the distance between the anchor and the positive while maximising the distance between the anchor and the negative through the triplet loss \cite{Triplet-Network}:
$$L = \max(d(a, p) - d(a, n) + \alpha, 0)$$

where \( d(x, y) \) is the distance between the embeddings of images \( x \) and \( y \), \( a \), \( p \), and \( n \) represent the anchor, positive, and negative images, respectively, and \( \alpha \) is a hyperparameter that defines how much the negative example should be farther away from the anchor compared to the positive. 

The implementation employs a custom semi-hard negative mining strategy to select informative negative samples. This approach accelerates learning without focusing on hard positives, which are associated with poor quality and mislabeled images.

Semi-hard negatives are instances that satisfy:
$d(a, n) > d(a, p)$ and $d(a, n) < d(a, p) + m$. It is applied in a batch as follows:

\begin{enumerate}
    \item For each anchor, find all positive and negative indices.
    \item Iterate over every positive pair, and apply semi-hard negative mining only from epoch 4 onwards to ensure stability.
    \begin{itemize}
        \item Select semi-hard negatives based on the above conditions.
        \item Apply inverse distance weighting to prioritise closer negatives.
        \item Randomly choose a negative index based on the weights.
    \end{itemize}
    \item For the first 3 epochs or if no semi-hard negatives are found, select a random negative.
    \item Compute triplet loss using the anchor, positive, and selected negative distances.
\end{enumerate}

Triplet Networks have largely been replaced by Angular Margin methods in facial recognition due to certain handicaps in the training process. As the number of individuals in the training set increases, the number of triplets grows polynomially, making it infeasible to compare every positive pair \cite{Arcface}, \cite{Cosface}. In this case, where the training set is orders of magnitude smaller than in facial recognition tasks, this does not pose an issue. In fact, the inability to select all the possible triplets for a training epoch may act as a regularisation mechanism, preventing the Triplet Network from rapidly overfitting to the training set.

\subsubsection{CosFace}

\begin{figure}[ht]
    \centering
    \includegraphics[width=1.0\textwidth]{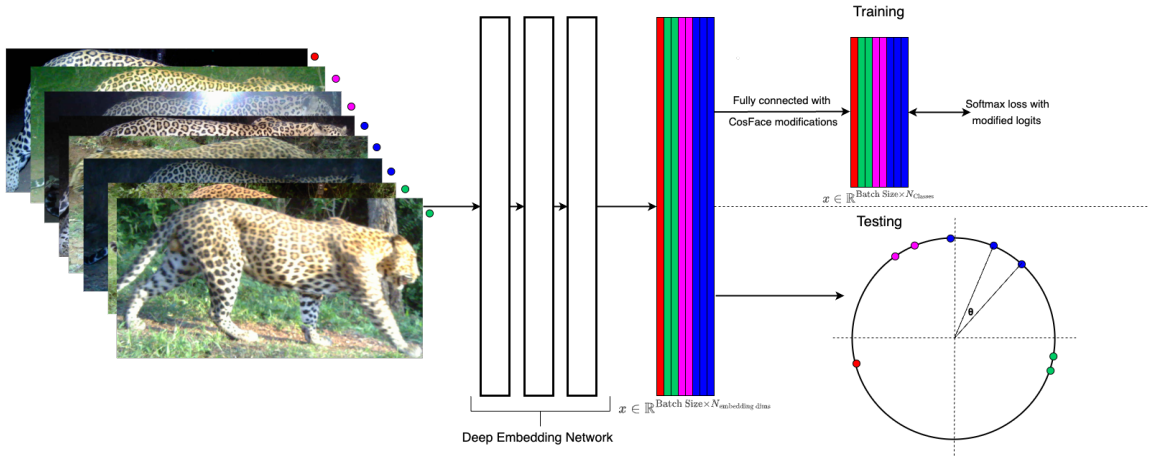}
    \caption{CosFace model representation. In the training phase, an extra fully-connected layer is added. The layer is discarded in testing, where embeddings are directly compared in the hypersphere through cosine similarity.}
    \label{fig:CosFace-architecture}
\end{figure}
Angular margin methods have become the state-of-the-art in facial recognition. Examples such as CosFace \cite{Cosface} combine the strengths of Triplet Networks and classification models while overcoming their respective weaknesses. Like Triplet Networks, these methods use a CNN to encode images as vector embeddings, enabling effective comparison. However, unlike Triplet Networks, angular margin methods can use information across an entire batch during training, similar to classification models \cite{Arcface}, \cite{Cosface}.

To comprehend the mechanics of CosFace, it is necessary to first examine the foundational concepts of the Softmax function, which converts logits into probabilities over $C$ classes and its associated loss.

\begin{equation*}
f(s)_i = \frac{e^{s_i}}{\sum_j^C e^{s_j}}
\end{equation*}

\begin{equation*}
L_{softmax} = -\frac{1}{N}\sum_i^N \log(f(s)_i)
\end{equation*}

Where $C$ is the number of classes, $N$ is the number of training samples, $s_i = W_{y_i}^{T}x_i + b_{y_i}$ is the logit of a fully connected layer corresponding to the $y_i$-th class and $x_i \in \mathbb{R}^d$ representing the embedding. It follows that the loss is minimal when: 

\begin{alignat*}{2}
\log(f(s)_i) &= 0 && \\
&\Longleftrightarrow && \frac{e^{s_i}}{\sum_j e^{s_j}} = 1 \\
&\Longleftrightarrow && e^{s_i} = \sum_j^C e^{s_j} \\
&\Longrightarrow && \sum_{j \neq i} e^{s_j} = 0
\end{alignat*}

Hence, this loss attempts to set the outputs for the incorrect classes to 0. However, this condition does not enforce any margin of separation between the vectors $x_i$ corresponding to each class, and only pushes $x_i$ to be similar to $W_{y_i}$. 

To introduce CosFace, it is essential to first reformulate the problem in angular space. By fixing the bias term $b_{y_i}$ to 0, then: $s_i = W_{y_i}^{T}x_i = \lVert W_{y_i} \rVert \lVert x_i \rVert \cos(\theta_i)$, with $\theta_i$ the angle between the weight vector and the exemplar vector. Given that in the testing and inference phase, cosine similarity is used to compare resemblance of two vectors, $\lVert W_{y_i} \rVert = 1$ and $\lVert x_i \rVert = s$ are fixed by L2 normalisation, thus fixing embedding vectors to a hypersphere of radius $s$. It thus follows that the normalised softmax loss can be written solely based on the angles:

\begin{equation*}
L_{N\_softmax} = -\frac{1}{N}\sum_i^N \log\frac{e^{s\cos{\theta_{y_i, i}}}}{\sum_j e^{s\cos{\theta_{j, i}}}}
\end{equation*}

It thus becomes clear that this loss attempts to minimise the angle between the weight vector and exemplar vector for correct class while maximising it for incorrect classes: $\cos{\theta_{y_i, i}} > \cos{\theta_{j, i}} \forall j \in C$. This condition is effective for closed-classification tasks where the classes seen in training and testing are the same, but it is not strict enough to learn a rich embedding space to apply to open-set learning like in wildlife PCR studies. Thus, \cite{Cosface} introduced a margin to make classes more separable in embedding space: $\cos{\theta_{y_i, i}} - m > \cos{\theta_{j, i}},  \forall j \in C, m \geq 0$. With this margin shrinking the effect of the correct class output, it enforces a much stronger separation of classes and thus more robust learning. The modified loss thus becomes:

\begin{equation*}
L_{CosFace} = -\frac{1}{N}\sum_i^N \log\frac{e^{s({\cos{\theta_{y_i, i}} - m})}}{e^{s(\cos{\theta_{y_i, i}} - m)} + \sum_{j \neq y_i}  e^{s\cos{\theta_{j, i}}}}
\end{equation*} 

To show the effectiveness of this loss, I built a CosFace model with a small subset of individual leopards, using just 3 embedding dimensions for visualisation in a normalised sphere. As can be seen below, when applying the CosFace loss, class separation is much less diffuse, which results in a better accuracy. 

\begin{figure}[ht]
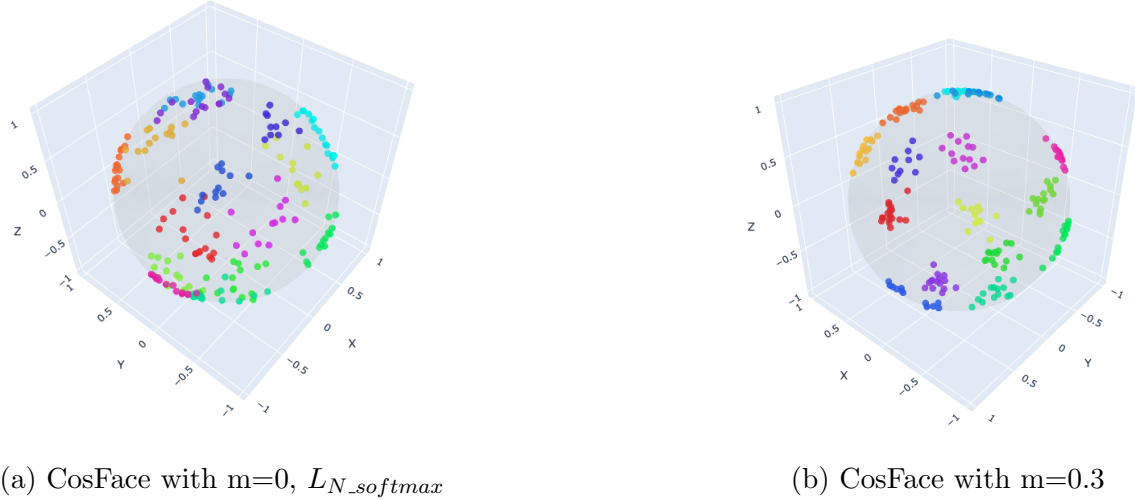

    \centering
    \begin{subfigure}[b]{0.4\textwidth}
        \includegraphics[width=\textwidth]{cosface_no_penalty.pdf}
        \caption{CosFace with m=0, $L_{N\_softmax}$}
        \label{fig:perfect}
    \end{subfigure}
    \hfill 
    \begin{subfigure}[b]{0.4\textwidth}
        \includegraphics[width=\textwidth]{cosface_penalty.pdf}
        \caption{CosFace with m=0.3}
    \end{subfigure}
    \caption{Showing how CosFace ensures class separation on a random selection of chosen leopards.}
\end{figure}
\label{fig:cosface_separation}

\subsubsection{Modified CosFace}
CosFace shows impressive results, however, certain aspects present opportunities for enhancement. Specifically, the margin $m$ applied is constant for every sample in the training. While this simplifies computation, it means that the same weighting is applied to every exemplar. Given that the loss is aggregated through the exemplars, the majority classes, those with numerous exemplars per ID, will dominate training. Although these majority classes are certainly informative, they pose a risk of overfitting. The model may cluster their embeddings by aligning them closely with the CosFace weights. This alignment, however, might occur without the model learning robust features.

To overcome this problem, \cite{modified-cosface} introduced a modified margin penalty which sets a higher penalty to exemplars less aligned with the weights. This approach encourages the model to concentrate on atypical, harder exemplars, promoting the development of robust features across exemplars. It also minimises the effect of majority classes dominating training, as the authors postulate that minority classes exhibit lower alignment with their weights.

Taking inspiration from this research and after heavy empirical testing, I introduce a novel CosFace modification margin function:

$$h(\theta) = \frac{\left(\left(1-\cos^{2}\left(x\right)\right)^{4}+0.1\right)}{1.1}$$

and the loss becomes:

\begin{equation*}
L_{\text{modified CosFace}} = -\frac{1}{N}\sum_i^N \log\frac{e^{s(\cos{\theta_{y_i, i}} - m \cdot h(\theta_{y_i, i}))}}{e^{s(\cos{\theta_{y_i, i}} - m \cdot h(\theta_{y_i, i}))} + \sum_{j \neq y_i}  e^{s\cos{\theta_{j, i}}}}
\end{equation*} 

The logic of the choice of function is as follows. Firstly, the margin function must be a function of $cos(\theta)$ and not $\theta$ due to the computational complexity of the $arccos$ function and its backpropagation. In this new function, the peak penalty is applied when vectors are orthogonal, representing the hard exemplars. Indeed, as vector dimensionality increases, the dot product of randomly chosen normalised vectors tends to approach zero, and as such they tend to orthogonality. Then, for exemplars that are angularly closer to the weight, a smaller penalty is applied to prevent them from dominating the training. Conversely, for exemplars with larger angular distances, which may represent lower-quality or misidentified images, the penalty is also reduced. 

\begin{figure}[ht]
    \centering
    \includegraphics[width=0.7\textwidth]{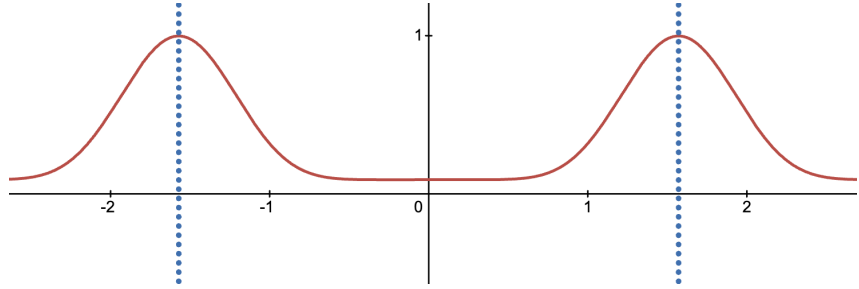}
    \caption{New margin function $h(\theta)$. Dotted lines at $\frac{-\pi}{2}$ and $\frac{\pi}{2}$}
    \label{fig:singleimage}
\end{figure}

\subsection{Interface}
While the core of this research focused on model optimisation, a basic user interface was developed to facilitate researchers to identify individual leopards. In facial recognition tasks, researchers often use nearest-neighbour approaches to cluster similar embeddings into individual IDs \cite{Triplet-Network}, \cite{KNN-cosface}. Given that this model's accuracy is still not close to perfect, using such a technique would yield numerous misidentifications. Due to this, a semi-automatic system was developed, balancing accuracy with efficiency.

The interface automatically passes input images through the preprocessing pipeline and through the best model. With the generated embeddings, the user is presented with the five most similar images to each new image, making an informed decision on whether each potential match is correct. This significantly reduces the visual workload from the traditional one-vs-all approach.

To add to this, a graph database is employed to store the matches. By utilising it, redundant comparisons are prevented and thus user effort is minimised. When a match between two images is confirmed, the database stores an edge. Then, when the comparison image is shown as anchor, the system excludes its known matches from the set of potential candidates presented to the user, preventing duplication of effort. The front-end implementation was assisted by generative AI tools.

\begin{figure}[ht]
    \centering
    \includegraphics[width=0.8\textwidth]{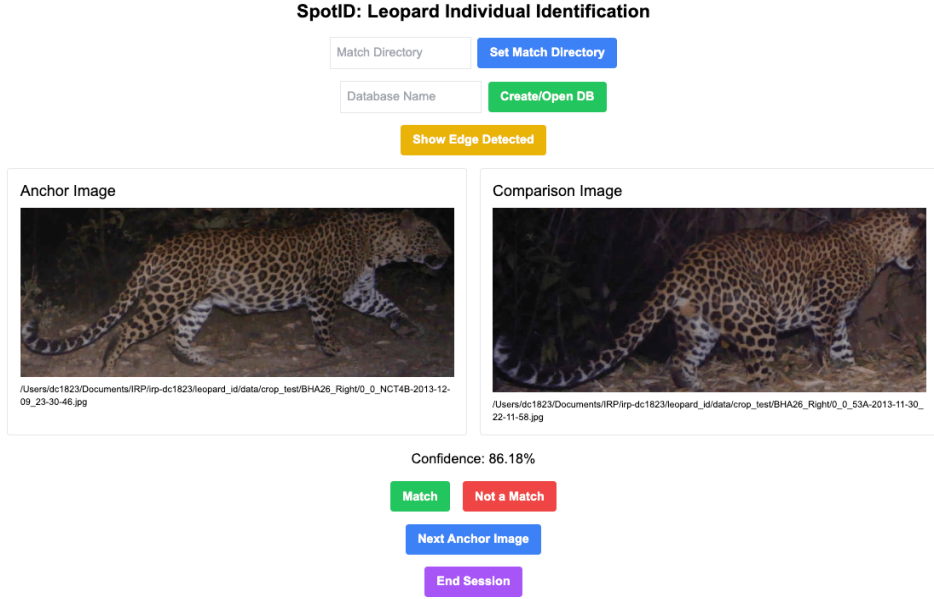}
    \caption{User Interface with a successful match of a leopard in different poses}
    \label{fig:singleimage}
\end{figure}

\subsection{Experimental Setup}
\subsubsection{Preprocessing and augmentation}
\label{subsec:Preprocessing}
Prior to model input, the images are passed through the preprocessing pipeline mentioned above, forming a 4 channel tensor. These are then resized to (512, 256). Random rotations of up to $10^\circ$ and a random crop of up to $10\%$ of each side are applied to the combined channels. A slight colour jitter augmentation is applied to the RGB channels. Finally, all channels are normalised. Augmentations are crucial given that leopards are always captured in varying poses and times of day, and a performance increase was seen after applying the augmentations.

\subsubsection{Sampling method}
In Triplet Networks, the choice of data sampler is paramount for effective training. Given that the number of unique leopards is large, randomly sampling exemplars in a batch would be ineffective, as the probability of choosing positive pairs to form triplets would be exceedingly small. I thus design a sampler that randomly chooses an individual leopard and loads $N$ random exemplars of that ID for that batch. After empirical testing, optimal $N$ was found to be 4. The same sampler was used for CosFace, as it resulted in more robust learning. Testing also revealed that sampling both left and right flanks within the same batch is more effective than sampling from only one side per batch. This approach enables the model to implicitly learn to distinguish between the flanks.

\subsubsection{Training}
The setup for Triplet and CosFace models is very similar. As demonstrated in figures ~\ref{fig:Triplet-architecture} and ~\ref{fig:CosFace-architecture}, both models require a backbone CNN to encode leopard images into embedding vectors. The Triplet Network acts directly on these embeddings, whereas CosFace adds one additional fully connected layer to the number of classes in the training set. This layer is then discarded for testing and inference. 

Given the lack of extensive data compared to facial recognition models, lightweight networks are essential to avoid overfitting. Both ResNet and EfficientNet architectures of similar sizes were tested as possible architectures. In my experiment with Triplet and CosFace models, an EfficientNetV2-B2 \cite{EfficientNet} demonstrated superior performance, yielding a relative improvement of 50-100\% in Dynamic Top-k Rank Average precision (which is defined in detail later) compared to ResNet-18 \cite{Resnet} consistently. These architectures possessed a similar number of parameters ($\approx 1 \times 10^7$). 

A fully-connected layer to an N-dimensional embedding space was added to both models. Lastly, the first layer was modified to accommodate the four-channel input tensor. After extensive empirical testing, it was seen that retraining the entire neural network yielded better results than fine-tuning the terminal layers of the model. Given that such a setup is more prone to overfitting \cite{transfer-learning}, data augmentation mentioned in subsection \ref{subsec:Preprocessing} was particularly crucial for effective training. 

The model was trained with an Adam optimiser \cite{adam}. For CosFace, I employed a learning rate scheduler starting at $\alpha=0.001$, while for the Triplet Network, I used a fixed learning rate of $\alpha=0.0008$. Batch sizes were smaller in Triplet Networks than in CosFace due to there being many more triplets than exemplars. They were thus fixed at 32 and 64 respectively. Lastly, the normalisation factor $s$ is set to 64.

93\% of the leopard flank ID's were used for training, corresponding to 1191 unique ID's and 7757 images. Of which, 309 ID's have just 1 image, which cannot be used to form positive pairs for the triplets. They are still sampled as possible negatives. In CosFace, these singleton ID's are dropped, as they add too many parameters to the CosFace fully connected layer and decrease the performance marginally.

\subsubsection{Testing}

88 unique flank ID's with 520 images were used for testing. In the context of a capture-recapture setting, verification of correctness needs a one-versus-all approach. This method involves comparing each exemplar vector against all others in the testing set using a suitable distance metric, Euclidean for Triplet Networks and Cosine for CosFace. The validity of the matches is then assessed by determining whether the most similar vectors correspond to correct matches. As such, the model performance is dependent on the number of unique flanks in the test set and is incomparable to the performance reported by other papers. With the similarity matrix for the whole dataset, two metrics can be defined:

\paragraph{Top-k Rank Match Detection (TkRMD)}
This metric is used in most wildlife PCR studies, as seen in \cite{sift1}, \cite{sift2}, \cite{sift3}, among others. In it, the top k most similar matches are examined for each query image. The proportion of cases that have at least one match in the top k most similar images is then measured, using only the flanks with at least 2 images per id. While this metric is informative, it has its flaws. Notably:
\begin{enumerate}
    \item \textbf{Insensitivity to multiple matches}: The metric only considers whether there is at least one correct match within the top-k results, which can lead to an overly optimistic assessment of the performance. Consequently, if only one correct match is present among the five most similar images for each image in the test set, this would result in a 100\% T5RMD, despite overlooking additional valid matches for individuals with more than two images.

    \item \textbf{Lack of Rank Sensitivity}: The metric treats all matches within the top-k equally. Hence, the metric attributes the same score for a model matching the leopard at rank 1 and a model matching it at rank k.
\end{enumerate}

\paragraph{Dynamic Top-k Rank Average Precision (DTkAP)}
To address the limitations of the above metric, I introduce this new evaluation measure. This metric is more stringent and provides a comprehensive assessment of the model's performance. Firstly, a dynamic top-k is defined: $k_i = \min(|C_i| - 1, k_{\max})$, where $C_i$ represents the number of exemplars in the class to which $i$ belongs and $k_{\max}$ an integer representing the maximum allowed k value. Then, I define:

$DTkAP(i) = \frac{1}{k_i} * \sum_{j=1}^{k_i} P(j)$, where $P(j)$ represents the precision at rank $j$. Hence, since the precision at each rank up to $k_i$ is calculated, it not only gives higher weighting for higher rankings, but it also rewards multiple matches. It is a more rigorous metric, and thus the absolute numbers need to be taken with caution. For example: If an image has a match at the 4th rank only, it would get 1 in the top-5 rank match detection, yet it would only get $\frac{1}{5} * (0+0+0+\frac{1}{4}+\frac{1}{5}) = 0.09$ in this metric.

\section{Results}
\label{sec:Results}

Two parallel training objectives were maximised to develop a robust and accurate leopard individual identification system:

\subsection{Preprocessing Pipeline Optimisation}

\begin{table}[h]
\centering
\begin{tabular}{|p{5cm}|c|c|}
\hline
\textbf{Pre-processing Pipeline} & \textbf{DT5AP} & \textbf{T5RMD} \\
\hline
Cropped RGB channels & 0.7815 & 0.9148 \\
\hline
Background removed RGB channels & 0.8486 & 0.9198 \\
\hline
Edge Detection channel & 0.6094 & 0.8034 \\
\hline
Background removed RGB channels + Edge Detection channel & 0.8284 & 0.9160 \\
\hline
Cropped RGB channels + Edge Detection channel & \textbf{0.8814} & \textbf{0.9533} \\
\hline
\end{tabular}
\caption{Comparing the performance of different pre-processing pipelines with the best performing modified CosFace parameters. Best performer in bold}
\label{tab:preprocessing}
\end{table}

The five preprocessing pipelines developed in section \ref{subsec: Preprocessing} were examined with the best model setup found in table \ref{tab:models}. The combination of cropped RGB channels with edge detection channel was found to yield the best performance. Interestingly, the cropped RGB channels alone produced worse performance than the background removed RGB channels, yet the combination of background removed with edge detection yielded a lower performance than the combination of cropped RGB channels and edge detection. I hypothesise that it could be due to the edge detected images deriving from the background removed images, and hence the errors in background removal and rosette extraction propagate.

Furthermore, edge detection alone performed considerably worse than the other methods. This observation is somewhat counterintuitive, given that edge detection isolates the leopard's rosette patterns. Several factors may affect this unexpected result. Firstly, EfficientNet was initially trained on RGB images, and hence the initial weights and architecture are tuned for such images. Furthermore, inconsistencies in rosette extraction for a subset of the data could have negatively influenced the training and testing process (see figure \ref{fig:edge detection}).

\subsection{Model and Hyperparameter optimisation}
\begin{table}[h]
\centering
\begin{tabular}{|p{5cm}|c|c|c|}
\hline
\textbf{Model} & \textbf{Parameters} & \textbf{DT5AP} & \textbf{T5RMD} \\
\hline
Naive model & - & 0.01976 & 0.09438 \\
\hline
HotSpotter & - & 0.9475 & 0.9654 \\
\hline
Best Triplet Model & dims=128, m=10 & 0.6907 & 0.8824 \\
\hline
Best Original CosFace & dims=1028, m=0.2 & 0.8413 & 0.9331 \\
\hline
Modified CosFace & dims=512, m=0.28 & 0.7025 & 0.8844 \\
\hline
Modified CosFace & dims=1028, m=0.1 & 0.8092 & 0.9210 \\
\hline
Modified CosFace & dims=1028, m=0.28 & \textbf{0.8814} & \textbf{0.9533} \\
\hline
Modified CosFace & dims=1028, m=0.4 & 0.7841 & 0.9225 \\
\hline
Modified CosFace & dims=2048, m=0.28 & 0.7866 & 0.9067 \\
\hline
\end{tabular}
\caption{Comparing the performance across models and the most important hyperparameters with the best preprocessing pipeline. Best performer of our models in bold.}
\label{tab:models}
\end{table}

Table \ref{tab:models} presents results for two key hyperparameters (embedding dimensions and margin), selected from numerous parameters tested during my comprehensive optimisation process. The modified CosFace model with dimensions=1028 and margin=0.28 achieved the best performance, slightly outperforming the original CosFace model in both metrics, demonstrating the utility of applying an adaptive margin to each exemplar rather than a fixed one.

CosFace and modified CosFace largely outperformed the Triplet Network baseline, suggesting a significant advance over previous research on patterned felid identification relying on Triplet Networks \cite{TripletTiger}. However, SIFT-based Hotspotter \cite{Hotspotter} still outperformed my best model, showing that there is still some room for improvement.

Notably, the modified CosFace implementation offers advantages other than the moderate performance improvement over the original implementation. My analysis showed that the test Class Cosine Distance Ratio (CCDR), defined as the ratio of intra-class to inter-class cosine distances, was considerably lower for the modified CosFace compared to the original (0.58 vs 0.65 respectively). This results in improved practical utility through higher difference in confidence values for real vs false matches in the user interface.

It's noteworthy that a considerably large embedding dimension increased performance. There's a trade-off to note: while smaller dimensions risk insufficient detail, larger spaces allow for more detailed representations at the cost of increased sparsity, hence why the 2048 dimensional embedding space proved to be too large for effective learning. Indeed, in the optimal 1024 dimensional space, even the closest images in the dataset were $23^{\circ}$ apart, compared to a staggering $87^{\circ}$ in a naive model.

Furthermore, experiments with margin sizes revealed that increasing the margin from 0.1 to 0.28 improved performance. Further increases led to decreased performance, suggesting potential overfitting. Indeed, for m=0.1 and m=0.4, the training set CCDRs were 0.35 and 0.27 respectively, showing tighter clustering with larger margins, as expected with figure \ref{fig:cosface_separation}. However, this improvement did not generalise to the test set, where CCDRs were 0.63 and 0.67 respectively, supporting the hypothesis of potential overfitting.

Finally, as shown below, T1RMD exceeds 90\% for the best model, significantly simplifying the effort of a user when verifying matches.

\begin{figure}[ht]
    \centering
    \includegraphics[width=0.8\textwidth]{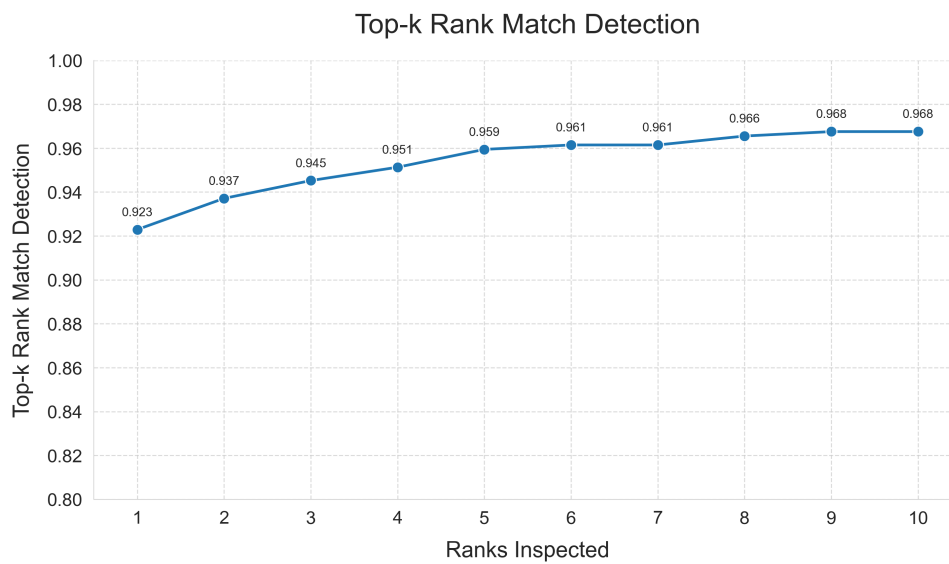}
    \caption{Top-k rank match detection on the test set inspecting up to 10 ranks for the best model. Note that the value differs slightly from the table due to this being the value at the final epoch, not the one with the highest DT5AP, which is the saved model.}
    \label{fig:singleimage}
\end{figure}

\newpage

\section{Discussion}
\label{sec:discussion}
\subsection{Preprocessing potential}

The extensive preprocessing employed in this research showed promising results. The addition of a Canny-edge-detection channel to the cropped image increased the performance of the model significantly. It highlighted the rosettes to guide the model without needless noise in the form of different illuminations and backgrounds. This approach has not been used in previous DL identification tasks and underlines the potential of using techniques that highlight the key features the model has to focus on while minimising information loss.

However, surprisingly, background removal did not yield the best performance. This could be attributed to the subset of images that had inaccurate background removal. With the recent rapid advances in segmentation models \cite{Segment}, investigating more robust background removal techniques could lead to advancements in animal individual identification. Additionally, exploring the generalisability of this preprocessing approach to other species with distinct patterns could discern further insights. 

\subsection{Challenges of limited dataset size}
\label{subsec:Challenges of limited dataset size}
One of the main challenges in applying DL to wildlife individual identification is the limited size of potential datasets. In facial recognition tasks, training datasets are not only of $10^6-10^7$ images, but also contain hundreds to thousands of images per individual \cite{Cosface}, \cite{modified-cosface}, \cite{Casia-Webface}. In contrast, the average number of images per leopard flank in this study was 6.4. This scarcity of data may impede the model's ability to learn inter-class separations that discriminate between the complex coat patterns of individual leopards. Indeed, in a high dimensional embedding space with limited data, distances become so vast that the model may easily overfit to separate the different IDs. With increased IDs, the sparse space becomes more densely populated, and thus more robust features must be learned to separate IDs effectively. To illustrate this, I tested the same model with just $40\%$ of the training IDs and obtained a significantly lower DT5AP of 0.5640 compared to the current model. This demonstrates the potential of increasing returns with larger datasets. To add to this, more extensive datasets would enable the testing of larger models such as an EfficientNetV2-B3 without such a high risk of overfitting.

With the Nature Conservation Foundation's constantly increasing database of leopard flanks, this research can be performed and improved yearly for better performances. However, for more endangered or elusive patterned wildlife, such as Snow Leopards, obtaining tens of thousands of tagged images is infeasible. In such cases, this research's outcome may not be a valid technique for their individual identification.

\subsection{Model Transferability and Dataset Considerations}
\label{subsec: Model transferability}

To evaluate the model's generalizability to other patterned species, we applied our trained leopard identification model to a subset of the Amur Tiger Re-identification in the Wild (ATRW) dataset \cite{TripletTiger}. Using 261 images of 18 individual tigers, the model achieved a DT5AP of 0.9200 and a T5RMD of 0.9883, surpassing its performance on the original leopard dataset. These results demonstrate two significant findings. 

Firstly, the model's ability to identify individual tigers without having been trained on this species suggests that the model's feature extraction is generalizable across patterned species, demonstrating robust transfer learning. This is particularly useful for application across a wide variety of species.

To add to this, the superior performance of the model on the ATRW dataset, compared to our leopard dataset is an indication that the data quality of the ATRW may be flawed. Indeed, as explained previously, most images of the ATRW dataset come from screenshots taken seconds apart. This means that the model may identify individual tigers due to the similar lightings and backgrounds, which may not reflect real-world identification scenarios such as the ones used in our leopard dataset. 

These findings demonstrate that this model may be applied directly to other patterned species if a large training dataset for that species is not found. If one such dataset is found, transfer learning from the leopard individual identification model as the initial weights should help the model converge swiftly. These results also underscore the importance of dataset diversity in computer vision models. While the ATRW dataset has been widely used as a benchmark, our analysis suggests that more challenging and diverse datasets may be necessary to accurately evaluate model performance for real-world applications.

\subsection{Comparability to SIFT-based techniques}

Hotspotter \cite{Hotspotter} is widely recognised as the state-of-the-art technique for wildlife PCR studies in recent years \cite{sift1}, \cite{sift2}, \cite{sift3}. It is the product of many years of fine tuning and perfecting, and as such, the accuracy is outstanding. In this dataset, Hotspotter is able to match images with high occlusion rates by identifying one rosette and matching it with other images. The DL model, on the other hand, struggles partially with occlusion of large parts of the leopard coat (see figure \ref{fig:twoimages_incorrect} for examples of wrong matches), although the introduction of augmentation techniques improved the performance with such images. This is in line with what has been seen in facial recognition tasks \cite{occlusion_facial_recognition}. 

It's worth noting that our dataset, originally compiled by the NCF using Wild-ID (a SIFT-based software like Hotspotter), may inadvertently favor Hotspotter's performance. This potential bias should be considered when interpreting the comparative results.

Although, as outlined above, this DL model has potential for improvement, further research may uncover that SIFT's explicit pattern matching might be better suited for more explicit patterns such as leopard rosettes. In the case of more subtle patterns like human faces, DL has already largely surpassed SIFT based techniques \cite{SIFT_worse_DL}. As such, one could employ the outcomes of this research and apply them to species with more subtle patterns, such as Kenyan Wild Dogs. Hotspotter has been shown to underperform with this species, with T1RMD as low as $44\%$ for a test dataset of comparable size to the one used in this study. The model's successful transfer to tigers above suggests that this could be done successfully, potentially offering a more versatile solution for wildlife identification tasks.

Thus, overall, while my modified CosFace has largely outperformed a Triplet Network baseline similar to those used in previous research with patterned species \cite{TripletTiger}, there is still room for improvement. With further calibrations, it may be possible to surpass the performance of Hotspotter, if not for leopard individual identification, then for other species with more subtle markings.

\section{Conclusion}

This study has advanced the field of wildlife identification through deep learning techniques, specifically for individual leopard recognition, with several key contributions: 

A novel adaptive margin CosFace model was designed. This model significantly surpasses the Triplet Network baseline and represents a modest improvement over the original CosFace implementation. The model achieved a robust performance, with a DT5AP of 0.8814 and T5RMD of 0.9533, only misidentifying exceedingly challenging samples, i.e. extremely rotated or highly occluded individuals. Furthermore, initial testing on other felid species (tigers) suggests promising cross-species applicability, highlighting the model's potential versatility.

The introduced preprocessing pipeline, which combines cropped RGB channels with edge detection, proved to be an effective technique to guide the model in learning important discriminative features from leopard coats. This approach could be adapted for use in other patterned species, broadening the scope of the research. 

While this method did not surpass the performance of the highly accurate SIFT-based Hotspotter, it represents a step forward in applying deep learning to patterned wildlife identification. The comparative analysis provided insights into the strengths and limitations of each technique. Although Hotspotter is more powerful in accurately identifying individuals with large occlusions, it has seen modest success in species with more subtle patterns. This limitation suggests that this research may have the potential to surpass Hotspotter by generalising to a wider range of pattern types, as seen in human facial recognition.

I have have demonstrated that modified angular margin methods in deep learning are a promising approach for PCR problems of patterned wildlife. The research contributes to both wildlife conservation and computer vision fields, offering a new pathway for future developments in wildlife monitoring.

\section{Data Accessibility}
The GitHub repository with the code for the project and interface can be found at \href{https://github.com/acse-dc1823/spotID}{https://github.com/acse-dc1823/spotID}. The training data is not available for the general public given the sensitive nature of it, although a minimum usable dataset for testing is provided with the repository. 

\section{Acknowledgements}
This project would not have been possible without the support and the data provided by the Nature Conservation Foundation and particularly Dr. Sanjay Gubbi. Hopefully, with this software, they can advance in their mission of monitoring and protecting the leopard in such a beautiful but fragmented habitat. I would also like to thank Dr. Lluís Guasch for his important insight and guidance throughout the project. I would lastly like to thank the Imperial HPC system and team for enabling me to train hundreds of test models.

\newpage
\appendix

\section*{Appendix}
\label{sec:Appendix}

\subsection*{Appendix A: Examples of suboptimal background removal and contrasting results of edge detection}
As explained in \ref{subsec: Preprocessing}, a small subset of images had incomplete background removals of usable leopard flanks, and again with the edge detection on the background removed. Some examples are shown here. 

\begin{figure}[htbp]
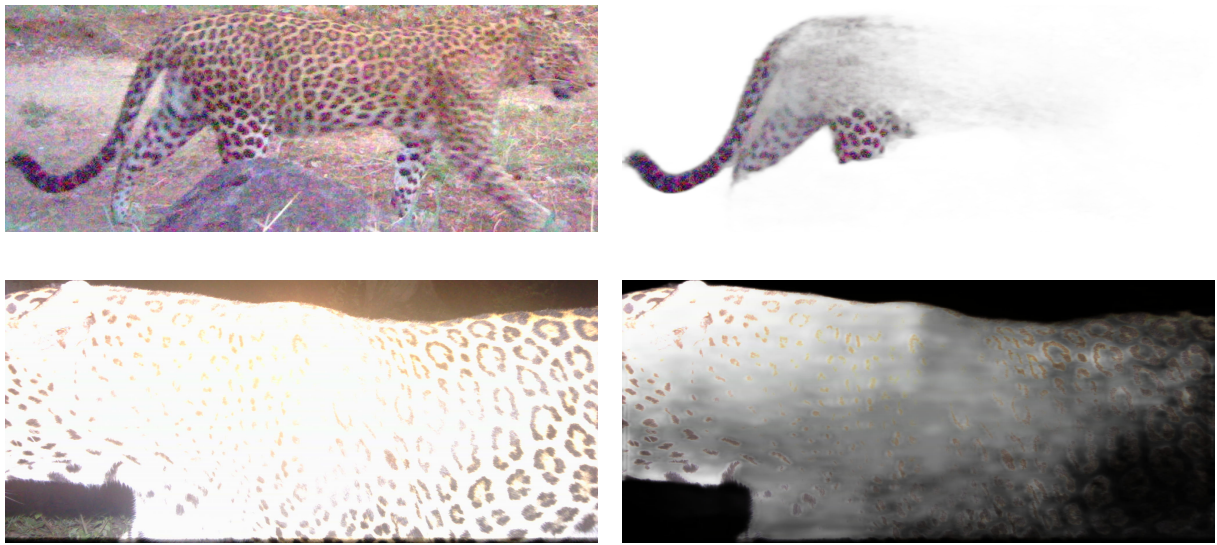

    \centering
    \begin{subfigure}[b]{0.49\linewidth}
        \centering
        \includegraphics[width=\textwidth]{cropped_image_bad_bg_rem.pdf}
        \label{fig:subfig_a}
    \end{subfigure}
    \hfill
    \begin{subfigure}[b]{0.49\linewidth}
        \centering
        \includegraphics[width=\textwidth]{bad_bg_rem1.pdf}
        \label{fig:subfig_b}
    \end{subfigure}
    
    \vspace{0.1cm}
    
    \begin{subfigure}[b]{0.49\linewidth}
        \centering
        \includegraphics[width=\textwidth]{cropped_image_bad_bg_rem_2.pdf}
        \label{fig:subfig_c}
    \end{subfigure}
    \hfill
    \begin{subfigure}[b]{0.49\linewidth}
        \centering
        \includegraphics[width=\textwidth]{bad_bg_rem2.pdf}
        \label{fig:subfig_d}
    \end{subfigure}
    \caption{Left: Cropped images. Right: Background removed images. Examples of usable leopard flanks where the background removal failed. Higher tendency of it happening in blurry or overexposed images.}
    \label{fig:bad background}
\end{figure}

\begin{figure}[htbp]
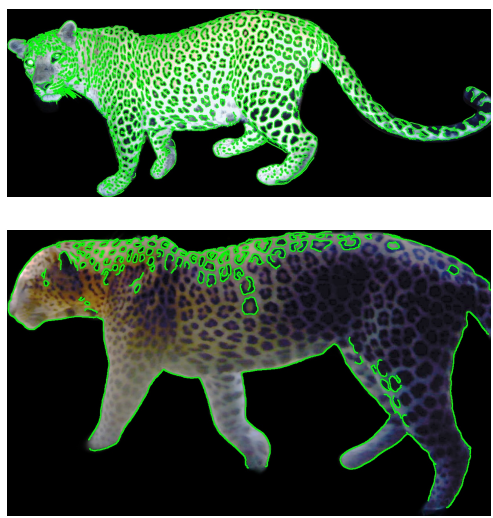

    \centering
    \begin{subfigure}[b]{0.4\textwidth}
        \centering
        \includegraphics[width=\textwidth]{0_0_BG-94A-2019-03-02_01-08-39-min.pdf}
    \end{subfigure}
    
    \vspace{1em}
    
    \begin{subfigure}[b]{0.4\textwidth}
        \centering
        \includegraphics[width=\textwidth]{0_0_BG-149B-2020-08-25_11-54-22.pdf}
    \end{subfigure}
    
    \caption{Edge detection output (green) joined with background removal for comparison. Top: Perfect edge detection output. Bottom: Incomplete extraction of rosettes.}
    \label{fig:edge detection}
\end{figure}

\newpage

\newpage

\subsection*{Appendix B: Contrasting Leopard Matching Outcomes and comparison to Hotspotter}

\begin{figure}[htbp]
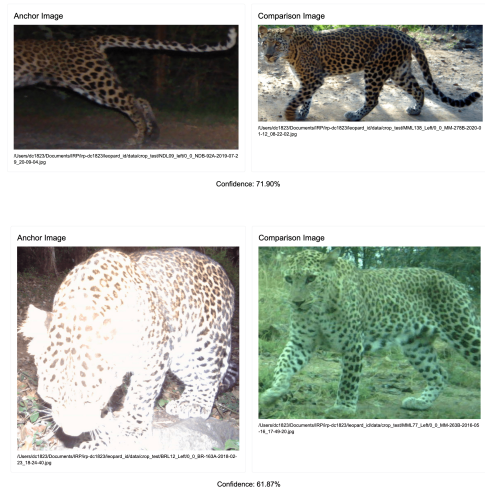

    \centering
    \begin{subfigure}[b]{0.4\textwidth}
        \centering
        \includegraphics[width=\textwidth]{matches/incorrect_1.pdf}
    \end{subfigure}
    
    \vspace{1em}
    
    \begin{subfigure}[b]{0.4\textwidth}
        \centering
        \includegraphics[width=\textwidth]{matches/incorrect_2.pdf}
    \end{subfigure}
    
    \caption{\small Examples of incorrect matches in the top 5 most similar images proposed by the model for leopards in the test set. High levels of occlusion and extreme rotation lower the accuracy.}
    \label{fig:twoimages_incorrect}
\end{figure}

\begin{figure}[htbp]
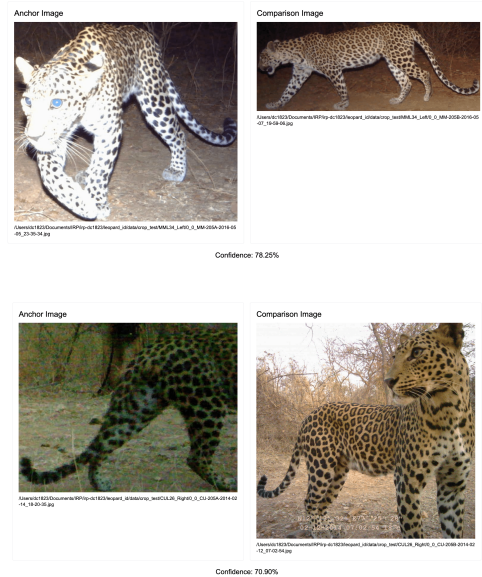

    \centering
    \begin{subfigure}[b]{0.4\textwidth}
        \centering
        \includegraphics[width=\textwidth]{matches/correct_1.pdf}
    \end{subfigure}
    
    \vspace{1em}
    
    \begin{subfigure}[b]{0.4\textwidth}
        \centering
        \includegraphics[width=\textwidth]{matches/correct_2.pdf}
    \end{subfigure}
    
    \caption{\small Examples of hard correct matches in the top 5 most similar images proposed by the model for leopards in the test set. Resistant to high levels of rotation, occlusion and overexposure.}
    \label{fig:twoimages_correct}
\end{figure}

\begin{figure}[htbp]
    \centering
    \begin{subfigure}[b]{0.3\textwidth}
        \centering
        \includegraphics[width=\textwidth]{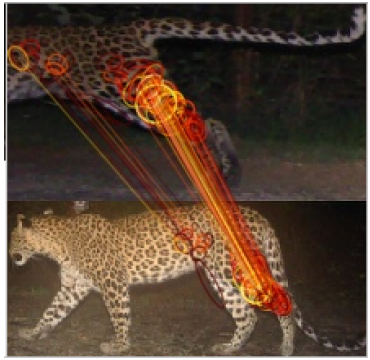}
    \end{subfigure}
    
    \vspace{1em}
    
    \begin{subfigure}[b]{0.3\textwidth}
        \centering
        \includegraphics[width=\textwidth]{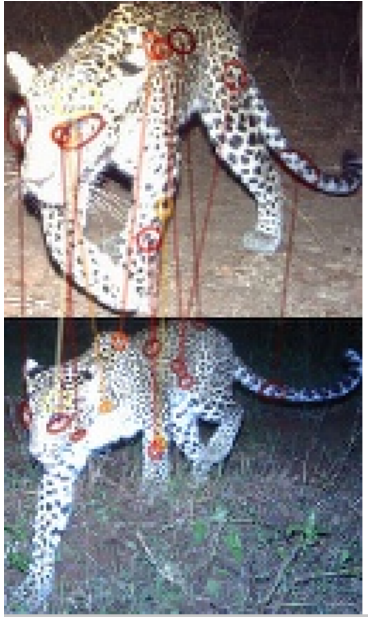}
    \end{subfigure}
    
    \caption{Top: correct match HotSpotter (was incorrect for CosFace model). Bottom: Incorrect match HotSpotter (was correct for CosFace model).}
    \label{fig:twoimages_correct}
\end{figure}

\newpage

\subsection*{Appendix C: Training Curves}

\begin{figure}[ht]
    \centering
    \includegraphics[width=0.8\textwidth]{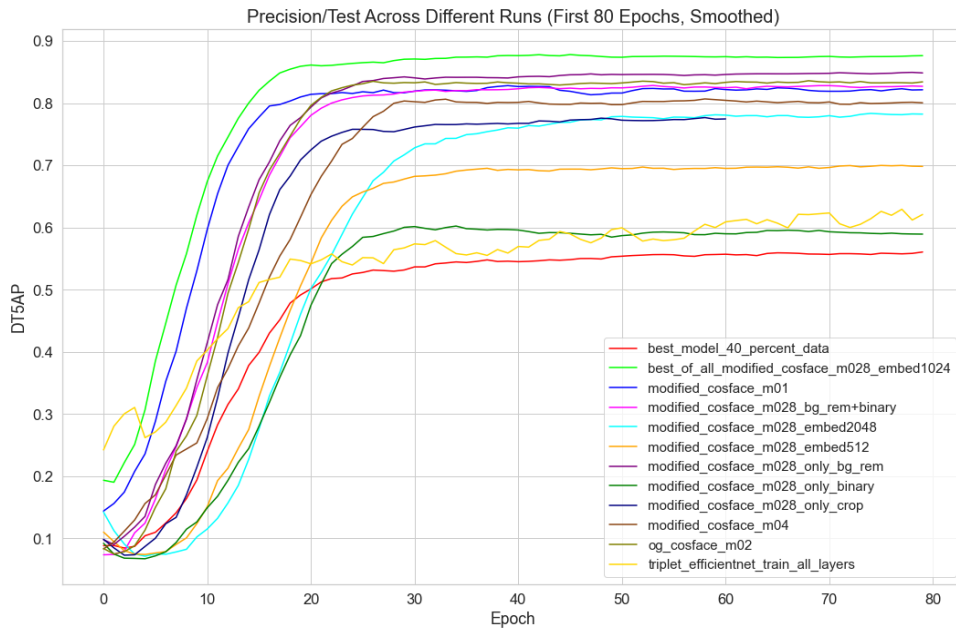}
    \caption{Testing DT5AP curves for models shown in section \ref{sec:Results} and mentioned in subsection \ref{subsec:Challenges of limited dataset size}. Triplet Network was trained up to 160 epochs, hence why the peak DT5AP seen here is lower than the one in Results.}
    \label{fig:DT5AP_curves}
\end{figure}

\subsection*{Appendix D: Tigers used for ATRW comparison}

The 18 randomly chosen tigers used to get the results for the ATRW comparison mentioned in \ref{subsec: Model transferability} are the following:

3, 13, 30, 42, 51, 64, 76, 97, 115, 126, 148, 178, 223, 243, 244, 252, 267, 274.

Taken from the train dataset (as it includes the tags).

\clearpage  


\end{document}